\documentclass[10pt,twocolumn,letterpaper]{article}

\usepackage[pagenumbers]{cvpr} %

\usepackage[normalem]{ulem}
\usepackage[dvipsnames]{xcolor}
\usepackage{stackengine}
\usepackage{xr}
\usepackage{amssymb}
\usepackage{pifont}
\usepackage{algorithm}
\usepackage{algpseudocode}%
\newcommand{\cmark}{\ding{51}}%
\newcommand{\xmark}{\ding{55}}%

\newcommand{\tablespace}{\vspace{-0.2cm}}
\usepackage{multirow}
\def\method{CausVid\xspace}
\usepackage{rotating}
\usepackage{setspace} %
\usepackage{adjustbox} %

\newboolean{arxiv}
\setboolean{arxiv}{false}

\makeatletter
\newcommand*{\addFileDependency}[1]{%
	\typeout{(#1)}
	\@addtofilelist{#1}
	\IfFileExists{#1}{}{\typeout{No file #1.}}
}
\makeatother

\definecolor{cvprblue}{rgb}{0.21,0.49,0.74}
\usepackage[pagebackref,breaklinks,colorlinks,citecolor=cvprblue]{hyperref}

\def\paperID{1409} %
\def\confName{CVPR}
\def\confYear{2025}

\title{From Slow Bidirectional to Fast Autoregressive Video Diffusion Models}

\author{Tianwei Yin$^{1^*}$ \hspace{6mm} Qiang Zhang$^{2^*}$ \hspace{6mm} Richard Zhang$^{2}$ \\ William T. Freeman$^{1}$ \hspace{6mm} Frédo Durand$^{1}$ \hspace{6mm} Eli Shechtman$^{2}$ \hspace{6mm} Xun Huang$^{2}$ \\ \vspace{-3mm} \\
$^{1}$MIT \hspace{5mm} $^{2}$Adobe \\\\
\centerline{\url{https://causvid.github.io/}}
}

\begin{document}

\twocolumn[{%
 \renewcommand\twocolumn[1][]{#1}%
 \maketitle
 \centering
 \includegraphics[width=\textwidth, trim={0 0.5cm 0 0.5cm}]{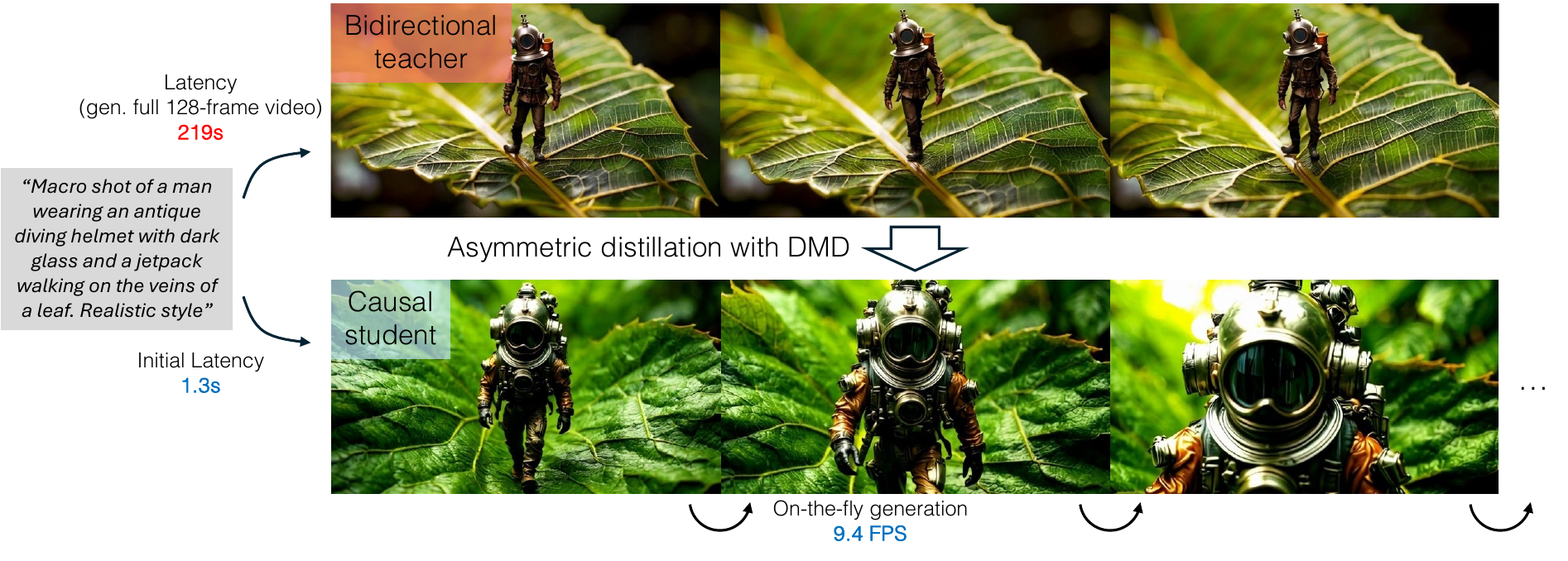}
 \captionof{figure}{ 
    Traditional bidirectional diffusion models (top) deliver high-quality outputs but suffer from significant latency, taking 219 seconds to generate a 128-frame video. Users must wait for the entire sequence to complete before viewing any results. In contrast, we distill the bidirectional diffusion model into a few-step autoregressive generator (bottom), dramatically reducing computational overhead. Our model~(\textbf{\method}) achieves an initial latency of only 1.3 seconds, after which frames are generated continuously in a streaming fashion at approximately 9.4 FPS, facilitating interactive workflows for video content creation. 
    \label{fig:teaser}
   }
 \vspace{1.8em}
}]

\maketitle

\let\thefootnote\relax\footnotetext{$^{*}$ Equal Contribution. Correspondence:~\texttt{tianweiy@mit.edu}
}

\begin{abstract}
    Current video diffusion models achieve impressive generation quality but struggle in interactive applications due to bidirectional attention dependencies.
    The generation of a single frame requires the model to process the entire sequence, including the future.
    We address this limitation by adapting a pretrained bidirectional diffusion transformer to an autoregressive transformer that generates frames on-the-fly.
    To further reduce latency, we extend distribution matching distillation (DMD) to videos, distilling 50-step diffusion model into a 4-step generator.
    To enable stable and high-quality distillation, we introduce a student initialization scheme based on teacher's ODE trajectories, as well as an asymmetric distillation strategy that supervises a causal student model with a bidirectional teacher.
    This approach effectively mitigates error accumulation in autoregressive generation, allowing long-duration video synthesis despite training on short clips.
    Our model achieves a total score of 84.27 on the VBench-Long benchmark, surpassing all previous video generation models.    
    It enables fast streaming generation of high-quality videos at 9.4 FPS on a single GPU thanks to KV caching. Our approach also enables streaming video-to-video translation, image-to-video, and dynamic prompting in a zero-shot manner.
    We release our code and pretrained models.
\end{abstract}

\vspace{-1em}

\begin{figure*}
    \centering
    \includegraphics[width=\linewidth]{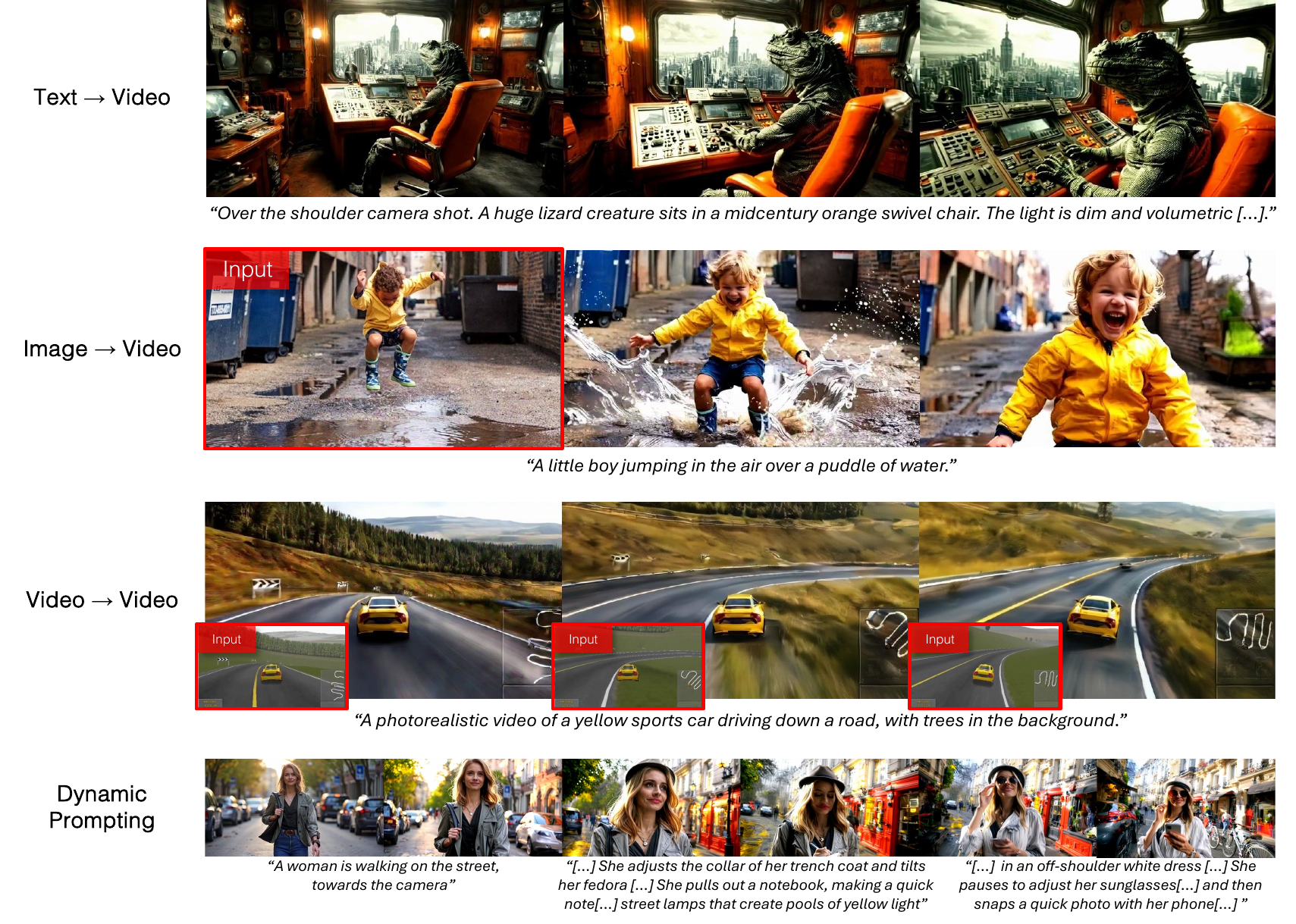}
    \caption{
    Diverse video generation tasks supported by our method. The model can generate videos from a single text prompt (top row) or with additional image input (second row). 
    Our model also enables interactive applications where generation results respond to user input with low latency.
    For example, it can add realistic texture and lighting to outputs rendered by a basic game engine that responds to user input on the fly (third row).
    Additionally, it enables dynamic prompting~(fourth row), allowing users to input new prompts at any point in a video to build extended narratives with evolving actions and environments.
    }
    \label{fig:main_result}
\end{figure*}

\section{Introduction}
\label{sec:intro}
The emergence of diffusion models has revolutionized how we can create videos from text~\cite{opensora,videoworldsimulators2024,jin2024pyramidal,yang2024cogvideox,polyak2024movie,ho2022video,blattmann2023stable}.
Many of the state-of-the-art video diffusion models rely on the Diffusion Transformer (DiT) architecture~\cite{peebles2023scalable,bao2022all}, which usually employs bidirectional attention across all video frames. 
Despite the impressive quality, the bidirectional dependencies imply that generating a single frame requires processing the entire video.
This introduces long latency and prevents the model from being applied to interactive and streaming applications, where the model needs to continually generate frames based on user inputs that may change over time. 
The generation of the current frame depends on future conditional inputs that are not yet available.
Current video diffusion models are also limited by their speed.
The compute and memory costs increase quadratically with the number of frames, which, combined with the large number of denoising steps during inference, makes generating long videos prohibitively slow and expensive.

Autoregressive models offer a promising solution to address some of these limitations, but they face challenges with error accumulation and computational efficiency.
Instead of generating all frames simultaneously, autoregressive video models generate frames sequentially. 
Users can start watching the video as soon as the first frame is generated, without waiting for the entire video to be completed. This reduces latency, removes limitations on video duration, and opens the door for interactive control.
However, autoregressive models are prone to error accumulation: each generated frame builds on potentially flawed previous frames, causing prediction errors to magnify and worsen over time.
Moreover, although the latency is reduced, existing autoregressive video models are still far from being able to generate realistic videos at interactive frame rate~\cite{jin2024pyramidal,kondratyuk2023videopoet,che2024gamegen}.

\begin{figure*}
    \centering
    \includegraphics[width=0.93\linewidth]{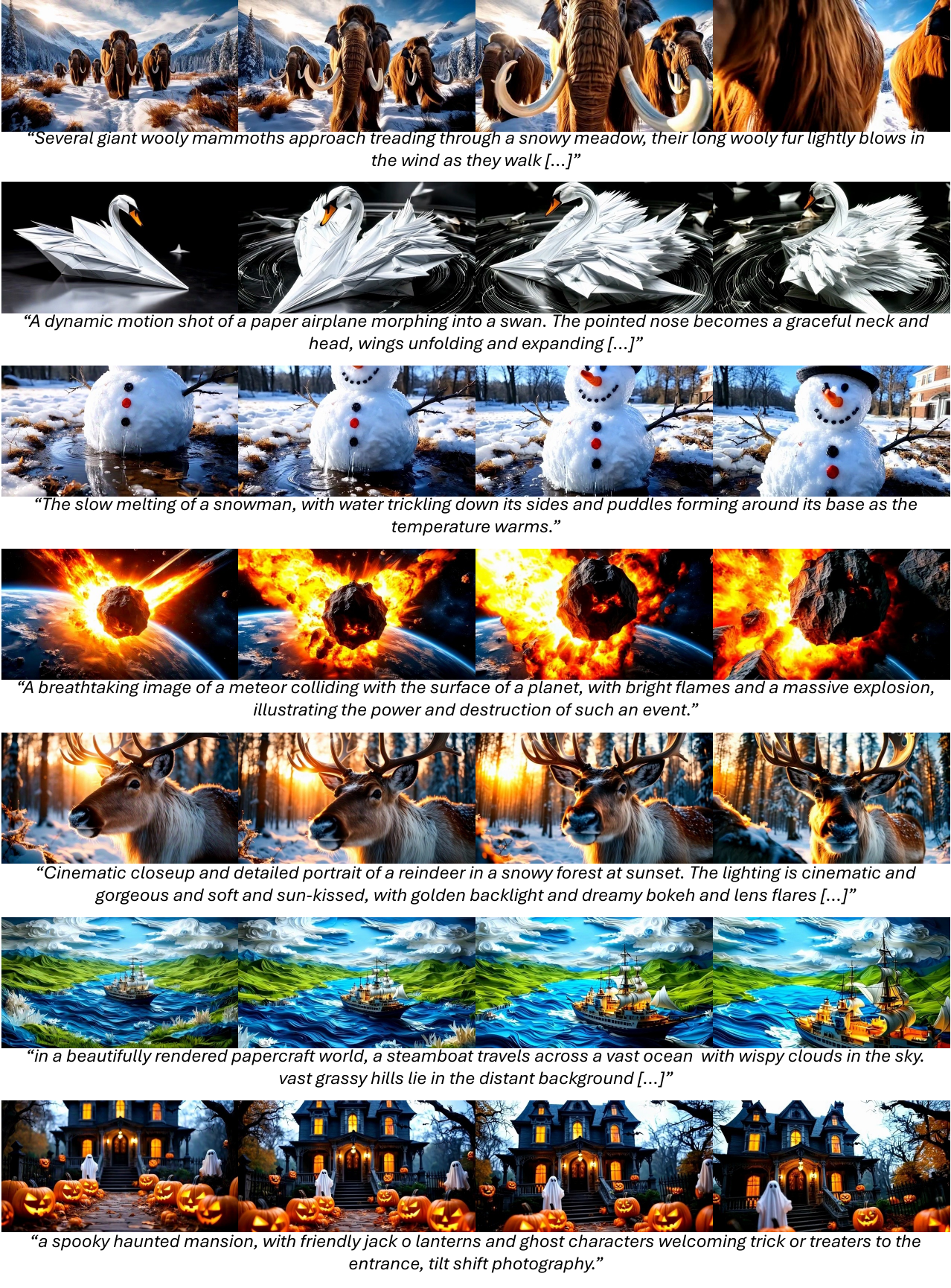}
    \vspace{-1em}
    \caption{
    Our model, CausVid, demonstrates that autoregressive video diffusion can be effectively scaled up for general text-to-video tasks, achieving quality on par with bidirectional diffusion models. Moreover, when combined with distillation techniques, it delivers multiple orders of magnitude speedup. Please visit our website for more visualizations.
    }
    \label{fig:t2v_extra}
\end{figure*}

\begin{figure*}
    \centering
    \includegraphics[width=\linewidth]{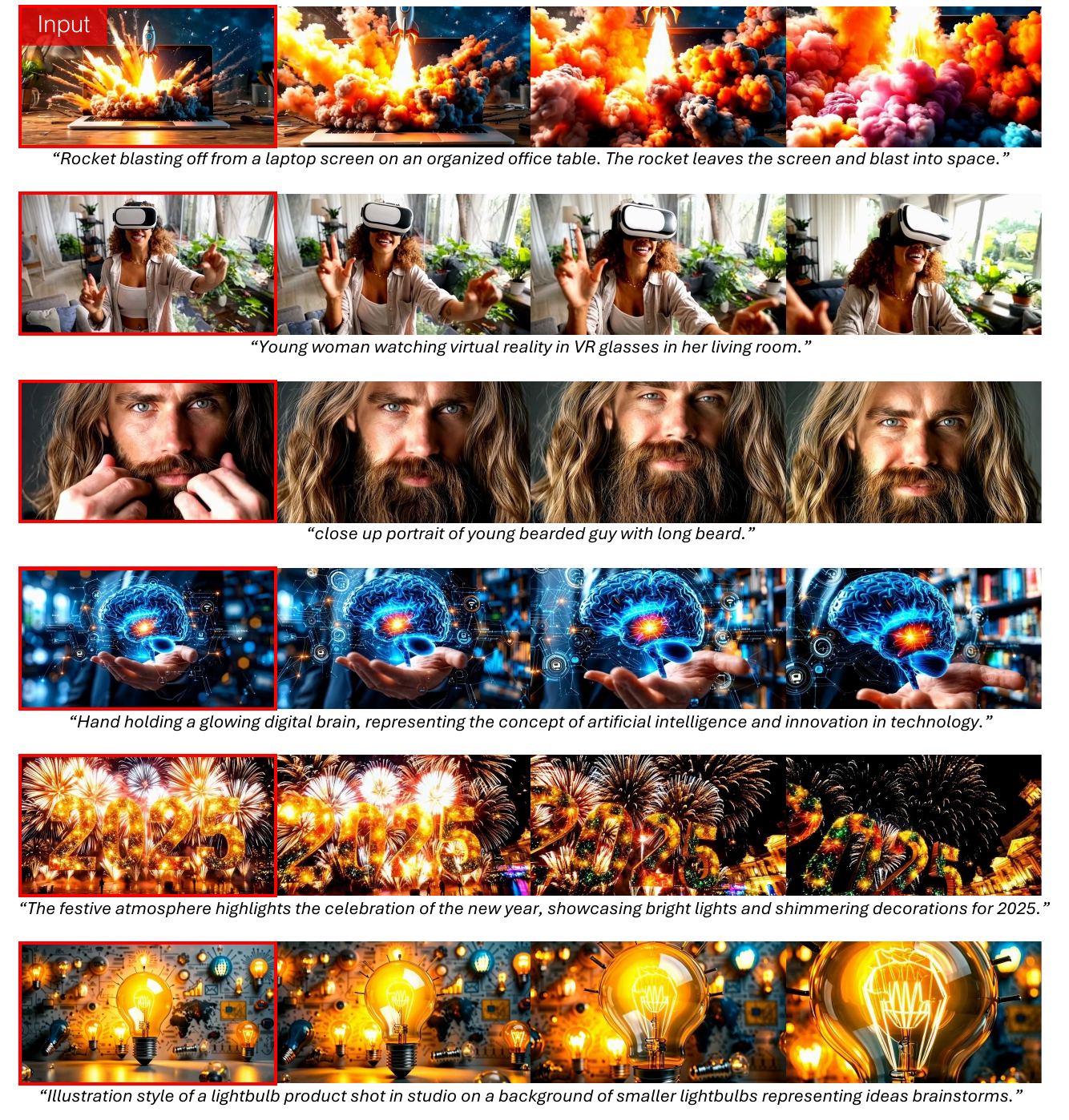}
    \vspace{-1em}
    \caption{
Trained exclusively on text-to-video generation, our model, CausVid, can be applied zero-shot to image-to-video tasks thanks to its autoregressive design. In the examples shown, the first column represents the input image, while the subsequent frames are generated outputs. Please visit our website for more visualizations.
    }
    \label{fig:i2v_extra}
\end{figure*}

In this paper, we introduce \textbf{\method}, a model designed for fast and interactive \textbf{caus}al \textbf{vid}eo generation.
We design an autoregressive diffusion transformer architecture with causal dependencies between video frames. Similar to the popular decoder-only large language models~(LLMs)~\cite{radford2019language,brown2020language}, our model achieves sample-efficient training by leveraging supervision from all input frames at each iteration, as well as efficient autoregressive inference through key-value (KV) caching. 
To further improve generation speed, we adapt distribution matching distillation~(DMD)~\cite{yin2024onestep,yin2024improved}, a few-step distillation approach originally designed for image diffusion models, to video data. 
Instead of naively distilling an autoregressive diffusion model~\cite{jin2024pyramidal,chen2024diffusion} into a few-step student, we propose an \textit{asymmetric distillation} strategy where we distill the knowledge in a pretrained teacher diffusion model with \textit{bidirectional} attention into our \textit{causal} student model. 
We show that this asymmetric distillation approach significantly reduced error accumulation during autoregressive inference.
This allows us to support autoregressively generating videos that are much longer than the ones seen during training. 
Comprehensive experiments demonstrate that our model achieves video quality on par with state-of-the-art bidirectional diffusion models while offering enhanced interactivity and speed. 
To our knowledge, this is the first autoregressive video generation method that competes with bidirectional diffusion in terms of quality~(Appendix Fig.~\ref{fig:t2v_extra} and Fig.~\ref{fig:i2v_extra}). 
Additionally, we showcase the versatility of our method in tasks such as image-to-video generation, video-to-video translation, and dynamic prompting, all achievable with extremely low latency~(Fig.~\ref{fig:main_result}).

\section{Related Work}
\label{sec:related}
\noindent\textbf{Autoregressive Video Generation.}
Given the inherent temporal order of video data, it is intuitively appealing to model video generation as an autoregressive process. 
Early research uses either regression loss~\cite{liu2017video,hao2018controllable} or GAN loss~\cite{mathieu2016deep,tulyakov2018mocogan,lee2018stochastic,vondrick2017generating} to supervise the frame prediction task. Inspired by the success of LLMs~\cite{brown2020language}, some works choose to tokenize video frames into discrete tokens and apply autoregressive transformers to generate tokens one by one~\cite{yan2021videogpt,liang2022nuwa,ge2022long,kondratyuk2023videopoet,wu2024vila,wang2024loong}. However, this approach is computationally expensive as each frame usually consists of thousands of tokens.
Recently, diffusion models have emerged as a promising approach for video generation. 
While most video diffusion models have bidirectional dependencies~\cite{opensora,videoworldsimulators2024,yang2024cogvideox,polyak2024movie}, autoregressive video generation using diffusion models has also been explored.
Some works~\cite{zhang2024extdm,valevski2024diffusion,alonso2024diffusion,jin2024pyramidal} train video diffusion models to denoise new frames given context frames.
Others~\cite{ruhe2024rolling,kim2024fifo,chen2024diffusion} train the model to denoise the entire video under the setting where different frames may have different noise levels. Therefore, they support autoregressive sampling as a special case where the current frame is noisier than previous ones.
A number of works have explored adapting pretrained text-to-image~\cite{kodaira2023streamdiffusion,liang2024looking,weng2024art,valevski2024diffusion} or text-to-video~\cite{henschel2024streamingt2v,xing2024live2diff,gao2024vid,xie2024progressive,kim2024fifo} diffusion models to be conditioned on context frames, enabling autoregressive video generation.
Our method is closely related to this line of work, with the difference that we introduce a novel adaption method through diffusion distillation, significantly improving efficiency and making autoregressive methods competitive with bidirectional diffusion for video generation.
\vspace{0.5em}

\noindent\textbf{Long Video Generation.}
Generating long and variable-length videos remains a challenging task.
Some works~\cite{qiu2023freenoise,chen2023seine,zhang2023diffcollage,zhang2024avid,wang2023gen,tan2024video,wang2024zola} generate multiple overlapped clips simultaneously using video diffusion model pretrained on fixed and limited-length clips, while employing various techniques to ensure temporal coherence.
Another approach is to generate long videos hierarchically, first generating sparse keyframes and then interpolating between them~\cite{yin2023nuwa,zhao2024moviedreamer}.
Unlike full-video diffusion models that are trained to generate videos of fixed length, autoregressive models~\cite{liang2022nuwa,henschel2024streamingt2v,jin2024pyramidal,gao2024vid,xie2024progressive,wang2024loong} are inherently suitable for generating videos of various length, although they may suffer from error accumulation when generating long sequences.
We find that the distribution matching objective with a bidirectional teacher is surprisingly helpful for reducing the accumulation of errors, enabling both efficient and high-quality long video generation.
\vspace{0.5em}

\noindent\textbf{Diffusion Distillation.} Diffusion models typically require many denoising steps to generate high-quality samples, which can be computationally expensive~\cite{ho2020denoising,song2020denoising}. 
Distillation techniques train a student model to generate samples in fewer steps by mimicking the behavior of a teacher diffusion model~\cite{salimans2022progressive,meng2023distillation,song2023consistency,yin2024onestep,xu2023ufogen,sauer2025adversarial,luhman2021knowledge,kang2024distilling,ren2024hyper}. 
Luhman~\etal~\cite{luhman2021knowledge} train a single-step student network to approximate the noise-image mapping obtained from a DDIM teacher model~\cite{song2020denoising}. Progressive Distillation~\cite{salimans2022progressive} trains a sequence of student models, reducing the number of steps by half at each stage. Consistency Distillation~\cite{song2023consistency,luo2023latent,kim2023consistency,song2024improved,heek2024multistep} trains the student to map any points on an ODE trajectory to its origin. Rectified flow~\cite{liu2022flow,liu2023instaflow,esser2024scaling} trains a student model on the linear interpolation path of noise-image pairs obtained from the teacher. Adversarial loss~\cite{goodfellow2020generative} is also used, sometimes in combination with other methods, to improve the quality of student output~\cite{sauer2025adversarial,lin2024sdxl,yin2024improved,kang2024distilling,xu2023ufogen}.
DMD~\cite{yin2024onestep,yin2024improved} optimizes an approximate reverse KL divergence~\cite{wang2023prolificdreamer,luo2023diff,franceschi2023unifying,yi2023monoflow}, whose gradients can be represented as the difference of two score functions trained on the data and generator's output distribution, respectively. 
Unlike trajectory-preserving methods~\cite{salimans2022progressive,kim2023consistency,liu2023instaflow}, DMD provides supervision at the distribution level and offers the unique advantage of allowing different architectural formulations for the teacher and student diffusion models. 
Our approach builds upon the effectiveness and flexibility of DMD to train an autoregressive generator by distilling from a bidirectional teacher diffusion model.

Recently, researchers have begun to apply distillation methods to video diffusion models, such as progressive distillation~\cite{lin2024animatediff}, consistency distillation~\cite{wang2023videolcm,wang2024animatelcm,mao2024osv, li2024t2v}, and adversarial distillation~\cite{mao2024osv,zhang2024sf}. 
Most approaches focus on distilling models designed to generate short videos~(less than 2 seconds).
Moreover, they focus on distilling a non-causal teacher into a student that is also non-causal. 
In contrast, our method distills a non-causal teacher into a causal student, enabling streaming video generation. 
Our generator is trained on 10-second videos and can generate infinitely long videos via sliding window inference.
There has been another line of work that focuses on improving the efficiency of video diffusion models by system-level optimization (e.g., caching and parallelism)~\cite{zhang2024cross,liu2024faster,zhao2024real,zou2024accelerating}. 
However, they are usually applied to standard multi-step diffusion models and can be combined with our distillation approach, further improving the throughput and latency.

\section{Background}
\label{sec:background}
This section provides background information on video diffusion models~(Sec.~\ref{sec:video_diffusion}) and distribution matching distillation~(Sec.~\ref{sec:dmd}), which our method is built upon.

\subsection{Video Diffusion Models}
\label{sec:video_diffusion}
Diffusion models~\cite{sohl2015deep,ho2020denoising} generate samples from a data distribution $p(x_{0})$ by progressively denoising samples that are initially drawn from a Gaussian distribution $p(x_{T})$. 
They are trained to denoise samples created by adding random noise $\epsilon$ to the samples $x_0$ from the data distribution
\begin{equation}
    x_t = \alpha_t x_0 + \sigma_t\epsilon,\ \epsilon \sim \mathcal{N}(0, I)\,,
    \label{eq:diffusion_noise}
\end{equation}
where $\alpha_t, \sigma_T>0$ are scalars that jointly define the signal-to-noise ratio according to a specific noise schedule~\cite{karras2022elucidating,kingma2021variational,song2020score} at step $t$. 
The denoiser with parameter $\theta$ is typically trained to predict the noise~\cite{ho2020denoising}
\begin{equation}
    \mathcal{L}(\theta) = \mathbb{E}_{t, x_0, \epsilon}\left\|\epsilon_{\theta}(x_t, t) - \epsilon\right\|_2^2\,.
    \label{eq:denoising_loss}
\end{equation}
Alternative prediction targets include the clean image $x_0$~\cite{salimans2022progressive,karras2022elucidating} or a weighted combination of $x_0$ and $\epsilon$ known as v-prediction~\cite{salimans2022progressive}.
All prediction schemes are fundamentally related to the score function, which represents the gradient of the log probability of the distribution~\cite{song2020score,kingma2021variational}: \begin{equation} \label{eq
} s_\theta(x_t, t) = \nabla_{x_t} \log p(x_t) = - \frac{\epsilon_\theta(x_t, t)}{\sigma_t}. \end{equation} In the following sections, we simplify our notation by using the score function $s_\theta$ as a general representation of the diffusion model, while noting that it can be derived through reparameterization from a pretrained model of any prediction scheme.
At inference time, we start from full Gaussian noise $x_T$ and progressively apply the diffusion model to generate a sequence of increasingly cleaner samples. There are many possible sampling methods~\cite{song2020denoising,lu2022dpm,zhangfast,karras2022elucidating} to compute the sample $x_{t-1}$ in the next time step from the current one $x_t$ based on the predicted noise $\epsilon_{\theta}(x_t, t)$.

Diffusion models can be trained on either raw data~\cite{ho2020denoising,jabri2022scalable,karras2022elucidating} or on a lower-dimensional latent space obtained by a variational autoencoder~(VAE)~\cite{rombach2022high,peebles2023scalable,karras2024analyzing,yang2024cogvideox,opensora}.
The latter is often referred to as latent diffusion models~(LDMs) and has become the standard approach for modeling high-dimensional data such as videos~\cite{zhou2022magicvideo,opensora,hong2022cogvideo,blattmann2023align,yang2024cogvideox}. 
The autoencoder usually compresses both spatial and temporal dimensions of the video, making diffusion models easier to learn. The denoiser network %
in video diffusion models can be instantiated by different neural architectures, such as U-Net~\cite{ronneberger2015u,ho2022video,zhou2022magicvideo,chen2023videocrafter1} or Transformers~\cite{vaswani2017attention,hong2022cogvideo,yang2024cogvideox,videoworldsimulators2024}.

\subsection{Distribution Matching Distillation}
\label{sec:dmd}
Distribution matching distillation is a technique designed to distill a slow, multi-step teacher diffusion model into an efficient few-step student model~\cite{yin2024improved, yin2024onestep}. 
The core idea is to minimize the reverse KL divergence across randomly sampled timesteps $t$ between the smoothed data distribution $p_{\text{data}}(x_{t})$ and the student generator's output distribution $p_{\text{gen}}(x_{t})$. 
The gradient of the reverse KL can be approximated as the difference between two score functions:
\begin{align}
\label{eq:dmd}
\nabla_{\phi} \mathcal{L}_{\text{DMD}} & \triangleq \mathbb{E}_{t} \left( \nabla_{\phi} \text{KL} \left( p_{\text{gen}, t} \| p_{\text{data}, t} \right) \right) \notag \\
& \approx - \mathbb{E}_{t} \left( \int \left( s_{\text{data}} \left( \Psi \left( G_{\phi}(\epsilon), t \right), t \right) \right. \right. \\ 
& \qquad \left. \left. - s_{\text{gen},\xi} \left( \Psi \left( G_{\phi}(\epsilon), t \right), t \right) \right) \frac{d G_{\phi}(\epsilon)}{d \phi} \, d\epsilon \right), \notag 
\label{eq:dmd}
\end{align}
where $\Psi$ represents the forward diffusion process as defined in Eq.~\ref{eq:diffusion_noise}, $\epsilon$ is random Gaussian noise, $G_{\phi}$ is the generator parameterized by $\phi$, and $s_{\text{data}}$ and $s_{\text{gen},\xi}$ represent the score functions trained on the data and generator's output distribution, respectively, using a denoising loss~(Eq.~\ref{eq:denoising_loss}). 

During training, DMD~\cite{yin2024onestep} initializes both score functions from a pre-trained diffusion model. 
The score function of the data distribution is frozen, while the score function of the generator distribution is trained online using the generator's outputs.
Simultaneously, the generator receives gradients to align its output with the data distribution~(Eq.~\ref{eq:dmd}).  
DMD2~\cite{yin2024improved} extends this framework from single-step to multi-step generation by replacing the pure random noise input $\epsilon$ with a partially denoised intermediate image $x_t$.

\begin{figure*}[ht!]
    \vspace{-30mm}
    \includegraphics[width=\textwidth]{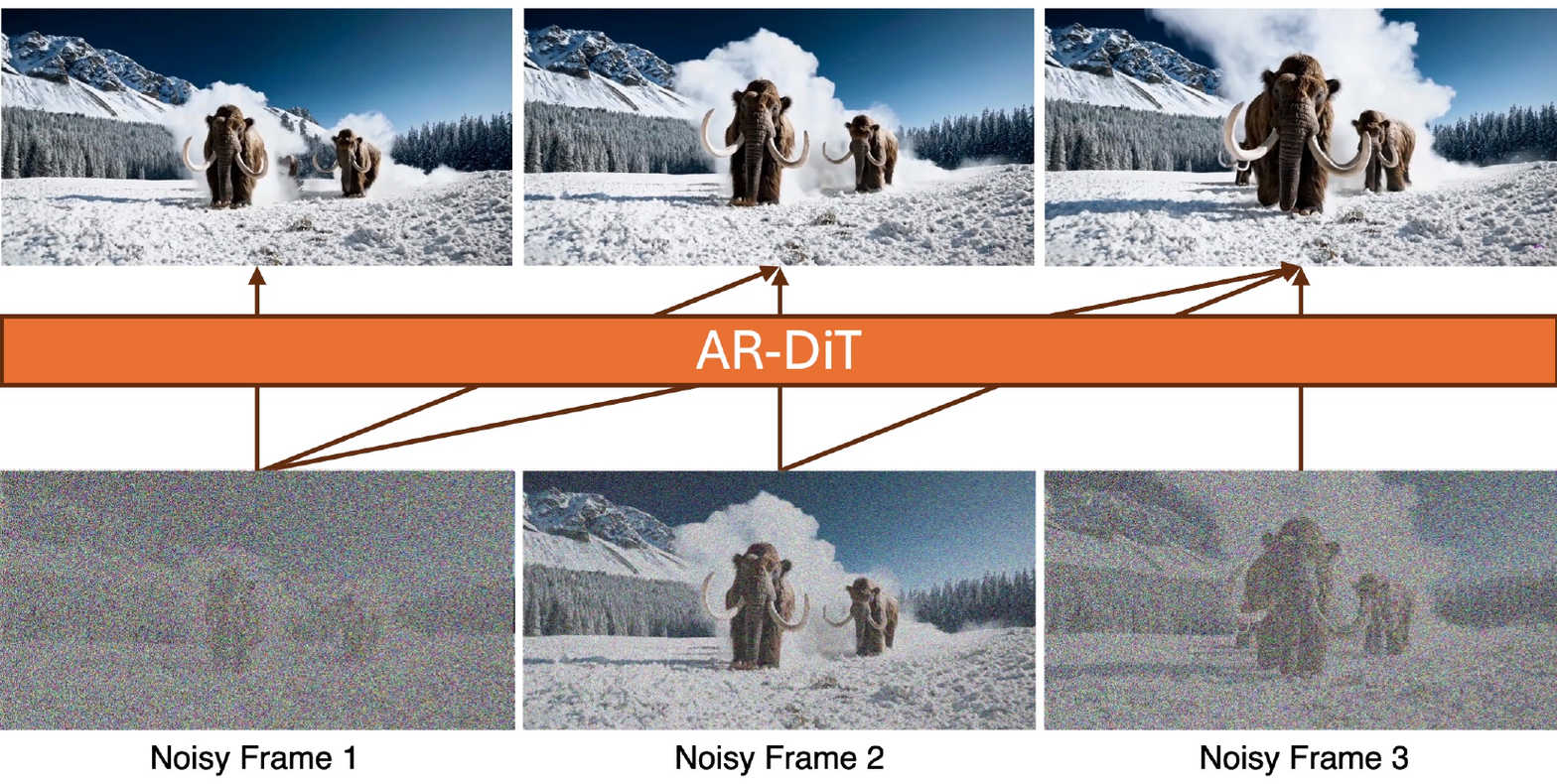}
    \vspace{-30mm}
    \caption{
    Overview of our autoregressive diffusion transformer architecture. Tokens in the current frame only attend to tokens from current and previous frames, but not the future.
    \label{fig:ardit} 
    }
\end{figure*}

\section{Methods}
\label{sec:method}

Our approach introduces an autoregressive diffusion transformer that enables sequential video generation~(Sec.~\ref{sec:autoregressive_arch}). 
We show our training procedure in Fig.~\ref{fig:method}, which uses asymmetric distillation~(Sec.~\ref{sec:teacher}) and ODE initialization~(Sec.~\ref{sec:student_init}) to achieve high generation quality and stable convergence. We achieve efficient streaming inference through KV caching mechanisms~(Sec.~\ref{sec:kv_cache}).

\subsection{Autoregressive Architecture}
\label{sec:autoregressive_arch}
We begin by compressing the video into a latent space using a 3D VAE. The VAE encoder processes each chunk of video frames independently, compressing them into shorter chunks of latent frames. The decoder then reconstructs the original video frames from each latent chunk.
Our causal diffusion transformer operates in this latent space, generating latent frames sequentially. We design a block-wise causal attention mechanism inspired by prior works that combine autoregressive models with diffusion~\cite{lee2023diffusion,zhou2024transfusion,liu2024autoregressive,li2024autoregressive}.
Within each chunk, we apply bidirectional self-attention among latent frames to capture local temporal dependencies and maintain consistency.
To enforce causality, we apply causal attention across chunks. This prevents frames in the current chunk from attending to frames in future chunks.
A visual illustration of the architecture of our autoregressive diffusion transformer is provided in Fig.~\ref{fig:ardit}. 
Our design maintains the same latency as fully causal attention, as the VAE decoder still requires at least a block of latent frames to generate pixels. 
Formally, we define the attention mask $M$ as
\begin{equation} 
M_{i,j} = 
\begin{cases} 
1, & \text{if } \left\lfloor \dfrac{j}{k} \right\rfloor \leq \left\lfloor \dfrac{i}{k} \right\rfloor, \\
0, & \text{otherwise.} 
\end{cases} 
\end{equation}
Here, $i$ and $j$ index the frames in the sequence, $k$ is the chunk size, and $\left\lfloor \cdot \right\rfloor$ denotes the floor function. 

\begin{figure*}
    \centering
    \includegraphics[width=\textwidth]{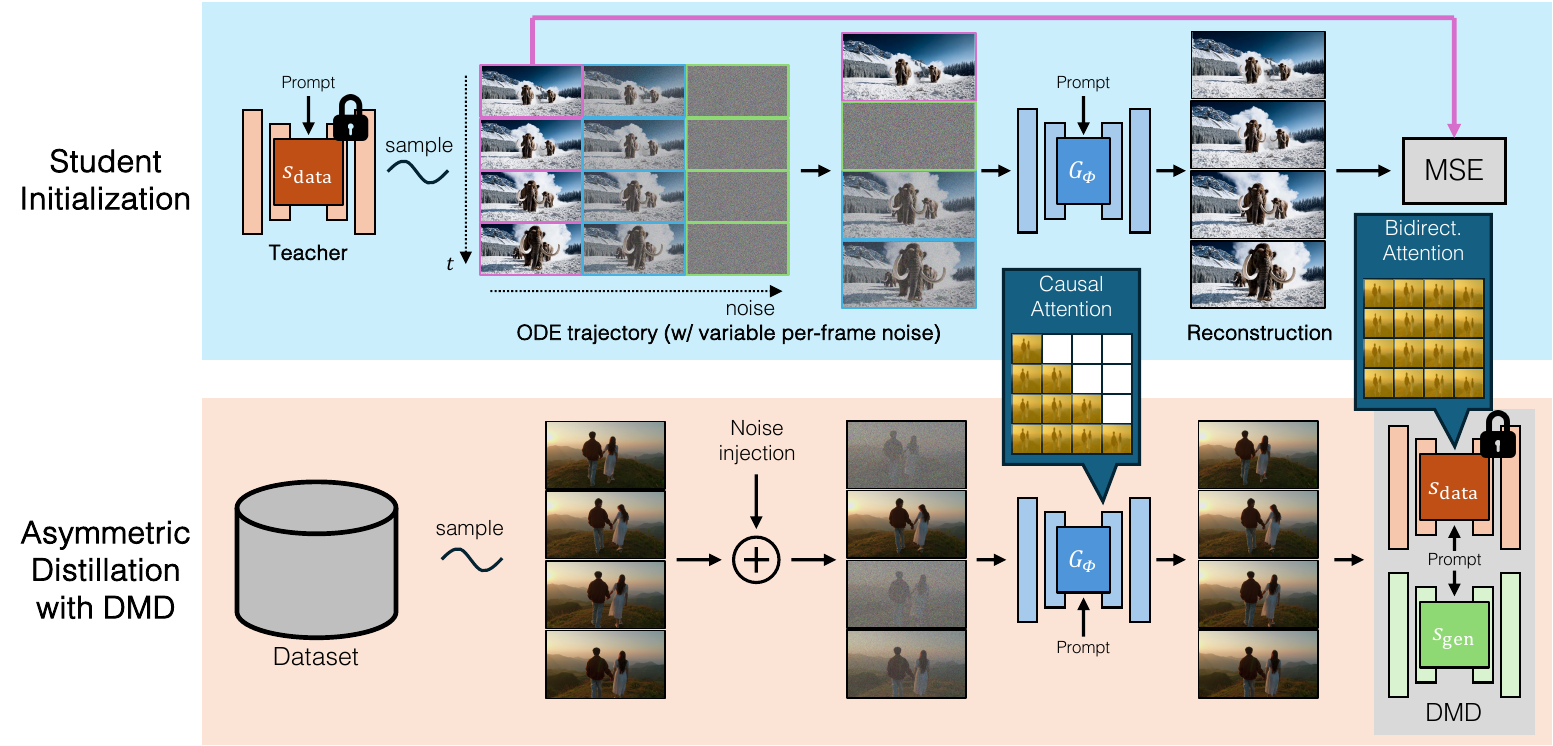}
    \caption{
    Our method distills a many-step, bidirectional video diffusion model $s_\text{data}$ into a 4-step, causal generator $G_\phi$. The training process consists of two stages. (top) Student Initialization: we initialize the causal student by pretraining it on a small set of ODE solution pairs generated by the bidirectional teacher (Sec.~\ref{sec:student_init}). This step helps stabilize the subsequent distillation training.
    (bottom) Asymmetric Distillation: using the \textit{bidirectional} teacher, we train the \textit{causal} student generator through a distribution matching distillation loss~(Sec.~\ref{sec:teacher}).
    \label{fig:method} 
    }\vspace{-2em}
\end{figure*}

Our diffusion model $G_{\phi}$ extends the DiT architecture~\cite{peebles2023scalable} for autoregressive video generation. 
We introduce block-wise causal attention masks to the self-attention layers (illustrated in Fig.~\ref{fig:method}) while preserving the core structure, allowing us to leverage pretrained bidirectional weights for faster convergence.

\subsection{Bidirectional \texorpdfstring{$\rightarrow$}{} Causal Generator Distillation}
\label{sec:teacher}

\begin{algorithm}[t]
\small 
\caption{Asymmetric Distillation with DMD \label{alg:training}}
\begin{algorithmic}[1]
\Require Few-step denoising timesteps $\mathcal{T} = \{0, t_1, t_2, \dots, t_Q \}$, video length $N$, chunk size $k$, pretrained bidirectional teacher model $s_{\text{data}}$, dataset $\mathcal{D}$.
\vspace{0.5em}
\State \textbf{Initialize} student model $G_\phi$ with ODE regression~(Sec.~\ref{sec:student_init})
\State \textbf{Initialize} generator output's score function $s_{\text{gen},\xi}$ with $s_{\text{data}}$

\While{training}
    \State \textbf{Sample} a video from the dataset and divide frames into $L = \lceil N / k \rceil$ chunks, $\{ x_0^{i} \}_{i=1}^L \sim \mathcal{D}$
    \State \textbf{Sample} per-chunk timesteps $\{ t^{i} \}_{i=1}^L\sim \text{Uniform}(\mathcal{T})$
    \State \textbf{Add noise:} ${ x_{t^i}^{i} = \alpha_{t^{i}} x_0^{i} + \sigma_{t^{i}} \epsilon^{i},\ \epsilon^{i} \sim \mathcal{N}(0, I)}$
    
    \State \textbf{Predict} clean frames with the student (block-wise causal mask): 
          $\hat{x}_0 = G_\phi\Bigl(\{ x_{t^i}^i \}, \{ t^i \}\Bigr)$\;

    \State \textbf{Sample} a single timestep $t\sim \text{Uniform}(0, T)$
    \State \textbf{Add noise} to predictions: $\hat{x}_t = \alpha_t \hat{x}_0 + \sigma_t \epsilon$, $\epsilon\sim \mathcal{N}(0, I)$
    
    \State \textbf{Update} student model $G_\phi$ using DMD loss $\mathcal{L}_{\text{DMD}}$ \Comment{Eq.~\ref{eq:dmd}}    
    
    \State \textbf{Train} generator's score function $s_{\text{gen},\xi}$:
    \State \quad \textbf{Sample new noise} $\epsilon'\sim \mathcal{N}(0, I)$
    \State \quad \textbf{Generate noisy} $\hat{x}_t$: $\hat{x}_t = \alpha_t \hat{x}_0 + \sigma_t \epsilon'$
    \State \quad Compute denoising loss and update $s_{\text{gen},\xi}$. \Comment{Eq.~\ref{eq:denoising_loss}}    
\EndWhile
\end{algorithmic}
\end{algorithm}

A straightforward approach to training a few-step causal generator would be through distillation from a causal teacher model. This involves adapting a pretrained bidirectional DiT model by incorporating causal attention mechanism described above and fine-tuning it using the denoising loss (Eq.~\ref{eq:denoising_loss}).
During training, the model takes as input a sequence of $N$ noisy video frames divided into $L$ chunks $\{x_t^{i}\}_{i=1}^{L}$, where $i \in \{1, 2, ..., L\}$ denotes the chunk index. Each chunk $x_t^{i}$ has its own noise time step $t^{i} \sim [0, 999]$, following Diffusion Forcing~\cite{chen2024diffusion}. 
During inference, the model denoises each chunk sequentially, conditioned on the previously generated clean chunks of frames.
While distilling this fine-tuned autoregressive diffusion teacher appears promising in theory, our initial experiments indicated that this naive approach yields suboptimal results. Since causal diffusion models typically underperform their bidirectional counterparts, training a student model from a weaker causal teacher inherently limits the student's capabilities. Moreover, issues such as error accumulation would propagate from teacher to student.
To overcome the limitations of a causal teacher, we propose an \textit{asymmetric distillation} approach: following state-of-the-art video models~\cite{videoworldsimulators2024,polyak2024movie}, we employ bidirectional attention in the teacher model while constraining the student model to causal attention~(Fig.~\ref{fig:method} bottom).
Algorithm~\ref{alg:training} details our training process.

\subsection{Student Initialization}
\label{sec:student_init}

Directly training the causal student model using the DMD loss can be unstable due to architectural differences. 
To address this, we introduce an efficient initialization strategy to stabilize training (Fig.~\ref{fig:method} top).

We create a small dataset of ODE solution pairs generated by the bidirectional teacher model:

\begin{itemize} 
    \item Sample a sequence of noise inputs $\{x_T^{i}\}_{i=1}^{L}$ from the standard Gaussian distribution $\mathcal{L}(0, I)$. 
    \item Simulate the reverse diffusion process with an ordinary differential equation (ODE) solver~\cite{song2020denoising} using the pretrained bidirectional teacher model to obtain the corresponding ODE trajectory $\{x_t^{i}\}_{i=1}^{L}$, where $t$ spans $T$ to $0$, covering all inference timesteps.  
\end{itemize}

\noindent 
From the ODE trajectories, we select a subset of $t$ values that match those used in our student generator. 
The student model is then trained on this dataset with a regression loss:
\begin{equation} \mathcal{L}_{\text{init}} = \mathbb{E}_{x,t^i}  || G_{\phi}(\{x^i_{t^i}\}_{i=1}^N, \{t^i\}_{i=1}^N) - \{x^i_0\}_{i=1}^N ||^2\,,
\end{equation}
where the few-step generator $G_{\phi}$ is initialized from the teacher model.
Our ODE initialization is computationally efficient, requiring only a small number of training iterations on relatively few ODE solution pairs. 

\subsection{Efficient Inference with KV Caching}
\label{sec:kv_cache}

During inference, we generate video frames sequentially using our autoregressive diffusion transformer with KV caching for efficient computation~\cite{brown2020language}. 
We show the detailed inference procedure in Algorithm~\ref{alg:inference}. 
Notably, because we employ KV caching, block-wise causal attention is no longer needed at inference time. 
This allows us to leverage a fast bidirectional attention implementation~\cite{dao2022flashattention}.

\begin{algorithm}[t]
\small 
\caption{\label{alg:inference}Inference Procedure with KV Caching} 
\begin{algorithmic}[1]
\Require Denoising timesteps $\{t_0=0, t_1, \dots, t_Q \}$, video length $N$, chunk size $k$, few-step autoregressive video generator $G_{\phi}$, 

\State Divide frames into $L = \lceil N / k \rceil$ chunks
\State \textbf{Initialize} KV cache $\text{C} \gets \emptyset$

\For{$i = 1$ to $L$}
    \State \textbf{Initialize current chunk:} $x_{t_Q}^{i} \sim \mathcal{N}(0, I)$
    
    \State \textbf{Iterative denoising over timesteps:}
    \For{$j = Q$ to $1$}
        \State \textbf{Generate output:} $\hat{x}_{t_j}^{i} = G_{\phi}(x_{t_j}^{i}, t_j)$ using cache $\text{C}$
        \State \textbf{Update chunk:} $x_{t_{j-1}}^{i} = \alpha_{t_{j-1}} \hat{x}_{t_j}^{i} + \sigma_{t_{j-1}} \epsilon'$ %
    \EndFor
    
    \State \textbf{Update KV cache:}
    \State Compute KV pairs with a forward pass $G_{\phi}(x_{0}^{i}, 0)$
    \State Append new KV pairs to cache $\text{C}$
\EndFor

\State \textbf{Return} $\{ x_{0}^{i} \}_{i=1}^{L}$
\end{algorithmic}
\end{algorithm}

\section{Experiments}
\label{sec:experiments}

\noindent\textbf{Models.}
Our teacher model is a bidirectional DiT~\cite{peebles2023scalable} with an architecture similar to CogVideoX~\cite{yang2024cogvideox}. The model is trained on the latent space produced by a 3D VAE that encodes $16$ video frames into a latent chunk consisting of $5$ latent frames. The model is trained on 10-second videos at a resolution of $352\times640$ and 12 FPS.
Our student model has the same architecture as the teacher, except that it employs causal attention where each token can only attend to other tokens within the same chunk and in preceding chunks.
Each chunk contains 5 latent frames. 
During inference, it generates one chunk at a time using 4 denoising steps, with inference timesteps uniformly sampled as $[999,748,502,247]$.
We use FlexAttention~\cite{he2024flexattention} for efficient attention computation during training.

\vspace{0.5em}
\noindent\textbf{Training.}
We distill our causal student model using a mixed set of image and video datasets following CogVideoX~\cite{yang2024cogvideox}.
Images and videos are filtered based on safety and aesthetic scores~\cite{schuhmann2022laion}.
All videos are resized and cropped to the training resolution~($352\times640$) and we use around 400K single-shot videos from an internal dataset, for which we have full copyright. 
During training, we first generate 1000 ODE pairs~(Sec.~\ref{sec:student_init}) and train the student model for $3000$ iterations with AdamW~\cite{loshchilov2017decoupled} optimizer and a learning rate of $5\times 10^{-6}$.
After that, we train with our asymmetric DMD loss~(Sec.~\ref{sec:teacher}) with the AdamW optimizer and a learning rate of $2\times 10^{-6}$ for $6000$ iterations.
We use a guidance scale of $3.5$ and adopt the two time-scale update rule from DMD2~\cite{yin2024improved} with a ratio of $5$.
The whole training process takes around 2 days on 64 H100 GPUs.

\vspace{0.5em}
\noindent\textbf{Evaluation.}
Our method is evaluated on VBench~\cite{huang2024vbench}, a benchmark for video generation with 16 metrics designed to systematically assess motion quality and semantic alignment.
For our main results, we use the first 128 prompts from MovieGen~\cite{polyak2024movie} to generate videos and evaluate model performance on three primary aspects from VBench competition’s evaluation suite\footnote{https://github.com/Vchitect/VBench/tree/master/competitions}. 
A comprehensive evaluation using all prompts from VBench is provided in the appendix.
The inference times are measured on a H100 GPU.

\subsection{Text to Video Generation}
We evaluate the ability of our approach to generate short videos~(5 to 10 seconds) and compare it against state-of-the-art methods: CogVideoX~\cite{yang2024cogvideox}, OpenSORA~\cite{opensora}, Pyramid Flow~\cite{jin2024pyramidal}, and MovieGen~\cite{polyak2024movie}. 
As shown in Tab.~\ref{table:short_result}, our method outperforms all baselines across all three key metrics: temporal quality, frame quality, and text alignment. Our model achieves the highest temporal quality score of $94.7$, indicating superior motion consistency and dynamic quality. 
In addition, our method shows notable improvements in frame quality and text alignment, scoring $64.4$ and $30.1$, respectively. 
In the supplementary material, we present our method’s performance on the VBench-Long leaderboard, achieving a total score of $84.27$ and securing first place among all officially evaluated video generation models.

\begin{figure}
    \centering
    \includegraphics[width=\linewidth]{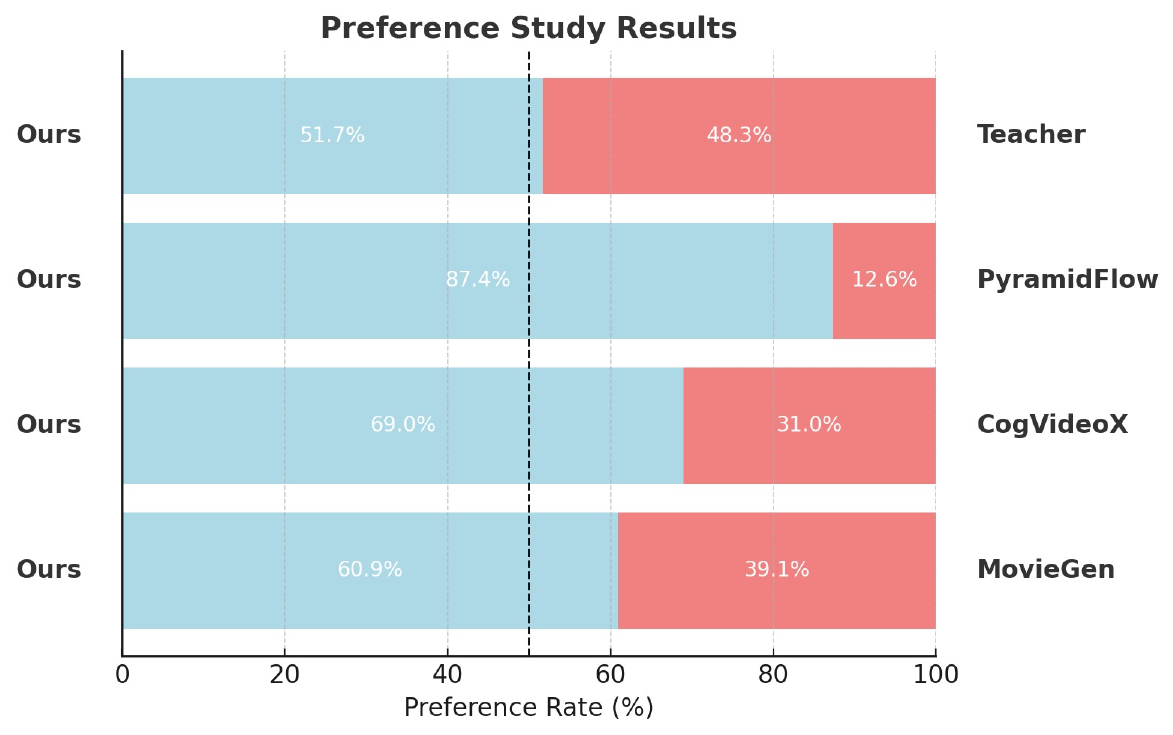}
    \caption{
    User study comparing our distilled causal video generator with its teacher model and existing video diffusion models. Our model demonstrates superior video quality (scores $>50\%$), while achieving a significant reduction in latency by multiple orders of magnitude.
    \label{fig:user_study} 
    }
\end{figure}

We further evaluate our model's performance through a human preference study. We select the first 29 prompts from the MovieGenBench dataset and collect ratings from independent evaluators using the Prolific platform. For each pair of compared models and each prompt, we collect 3 ratings from different evaluators, resulting in a total of 87 ratings per model pair. The evaluators choose the better video between the two generated videos based on the visual quality and semantic alignment with the input prompt.
The specific questions and interface are shown in Appendix Fig.~\ref{fig:interface}. For reproducibility, we use a fixed random seed of zero for all videos.
As illustrated in Fig.~\ref{fig:user_study}, our model consistently outperforms baseline methods such as MovieGen, CogVideoX, and Pyramid Flow. Notably, our distilled model maintains performance comparable to the bidirectional teacher while offering orders of magnitude faster inference, validating the effectiveness of our approach.

\begin{figure}
    \centering
    \includegraphics[width=\linewidth, trim={0 0.8cm 0 1.1cm}]{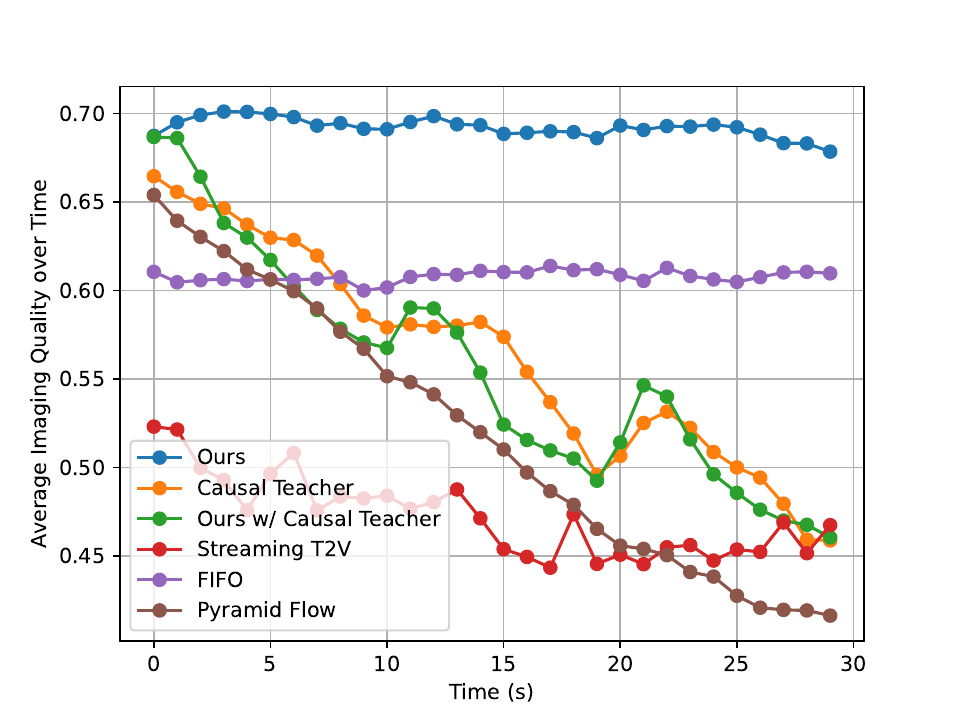}
    \caption{Imaging quality scores of generated videos over 30 seconds. Our distilled model and FIFO-Diffusion are the most effective at maintaining imaging quality over time. 
    The sudden increase of score for the causal teacher around 20s is due to a switch of the sliding window, resulting in a temporary improvement in quality. %
    }
    \label{fig:degradation}
\end{figure}

We also compare our method with prior works designed for long video generation: Gen-L-Video~\cite{wang2023gen}, FreeNoise~\cite{qiu2023freenoise}, StreamingT2V~\cite{henschel2024streamingt2v}, FIFO-Diffusion~\cite{kim2024fifo}, and Pyramid Flow~\cite{jin2024pyramidal}. 
We use a sliding window inference strategy, taking the final frames of the previous 10-second segment as context for generating the next segment.
The same strategy is also applied to generate long videos using Pyramid Flow. 
Tab.~\ref{table:long_result} shows that our method outperforms all baselines in terms of temporal quality and frame-wise quality and is competitive in text alignment.
It also successfully prevents error accumulation.
As shown in Fig.~\ref{fig:degradation}, our method maintains image quality over time, while most autoregressive baselines suffer from quality degradation~\cite{henschel2024streamingt2v,jin2024pyramidal,chen2024diffusion}.

Tab.~\ref{table:efficiency} compares the efficiency of our method with competing approaches~\cite{yang2024cogvideox,jin2024pyramidal} and our bidirectional teacher diffusion model.
Our method achieves a $160\times$ reduction in latency and a $16\times$ improvement in throughput compared to the similarly scaled CogVideoX~\cite{yang2024cogvideox}.

\begin{table}[!ht]
\centering
\scriptsize
\begin{tabular}{l@{\ \ \ }c@{\ \ \ }c@{\ \ \ }c@{\ \ \ }c}
\toprule
\multirow{2}{*}{Method} & \multirow{2}{*}{Length~(s)} & \multirow{2}{*}{\shortstack[c]{Temporal \\ Quality}} & \multirow{2}{*}{\shortstack[c]{Frame \\ Quality}} & \multirow{2}{*}{\shortstack[c]{Text \\ Alignment}} \\
\\
\midrule 
CogVideoX-5B  & 6 &  89.9 & 59.8 & 29.1   \\ 
OpenSORA & 8 & 88.4 & 52.0 & 28.4  \\ 
Pyramid Flow & 10 & 89.6 & 55.9 & 27.1 \\
MovieGen & 10 & 91.5 & 61.1 & 28.8\\
\textbf{\method~(Ours)} & 10 &\textbf{94.7} & \textbf{64.4} & \textbf{30.1} \\ 
\bottomrule
\end{tabular}
\tablespace
\caption{\label{table:short_result} Evaluation of text-to-short-video generation. Each method is evaluated at its closest supported length to 10 seconds.}
\end{table}

\begin{table}[h!]
\centering
\scriptsize
\begin{tabular}{l@{\ \ \ }c@{\ \ \ }c@{\ \ \ }c}
\toprule
\multirow{2}{*}{Method} & \multirow{2}{*}{\shortstack[c]{Temporal \\ Quality}} & \multirow{2}{*}{\shortstack[c]{Frame \\ Quality}} & \multirow{2}{*}{\shortstack[c]{Text \\ Alignment}} \\
\\
\midrule 
Gen-L-Video  & 86.7 & 52.3 & 28.7\\ 
FreeNoise  & 86.2 & 54.8 & 28.7 \\ 
StreamingT2V   & 89.2 & 46.1 & 27.2  \\ 
FIFO-Diffusion  & 93.1 & 57.9 & \textbf{29.9} \\ 
Pyramid Flow & 89.0 & 48.3 & 24.4    \\
\textbf{\method~(Ours)} & \textbf{94.9} & \textbf{63.4} & 28.9  \\
\bottomrule
\end{tabular}
\tablespace
\caption{\label{table:long_result} Evaluation of text-to-long-video generation. All methods produce videos approximately 30s in length.}
\end{table}

\begin{table}[h!]
\centering
\scriptsize
\begin{tabular}{lrc}
\toprule
Method & Latency~(s) & Throughput~(FPS)\\
\midrule 
CogVideoX-5B & 208.6 & 0.6 \\ 
Pyramid Flow & 6.7 & 2.5 \\
Bidirectional Teacher & 219.2 & 0.6 \\
\textbf{\method~(Ours)} & \textbf{1.3} & \textbf{9.4}\\
\bottomrule
\end{tabular}
\tablespace
\caption{\label{table:efficiency} Latency and throughput comparison across different methods for generating 10-second, 120-frame videos at a resolution of $640\times352$. The total time includes processing by the text encoder, diffusion model, and VAE decoder. Lower latency ($\downarrow$) and higher throughput ($\uparrow$) are preferred.}
\end{table}

\subsection{Ablation Studies}
\label{sec:ablation}
\definecolor{myorange}{RGB}{255,127,13}
\definecolor{mygreen}{RGB}{45,160,43}
\definecolor{myblue}{RGB}{31,119,180}
First, we present results of directly fine-tuning the bidirectional DiT into a causal model, without using few-step distillation. We apply causal attention masks to the model and fine-tune it with the autoregressive training method described in Sec.~\ref{sec:teacher}.
As shown in Tab.~\ref{table:ablation}, the many-step causal model performs substantially worse than the original bidirectional model. 
We observe that the causal baseline suffers from error accumulation, leading to rapid degradation in generation quality over time~(\textcolor{myorange}{orange} in Fig.~\ref{fig:degradation} ).

We then conduct an ablation study on our distillation framework, examining the student initialization scheme and the choice of teacher model. Tab.~\ref{table:ablation} shows that given the same ODE initialization scheme~(as introduced in Sec.~\ref{sec:student_init}), the bidirectional teacher model outperforms the causal teacher model and is also much better than the initial ODE-fitted model~(where we denote the teacher as None). 
As shown in Fig.~\ref{fig:degradation}, the causal diffusion teacher suffers from significant error accumulation~(\textcolor{myorange}{orange}), which is then transferred to the student model~(\textcolor{mygreen}{green}). 
In contrast, we find that our causal student model trained with our asymmetric DMD loss and a bidirectional teacher~(\textcolor{myblue}{blue}) performs much better than the many-step causal diffusion model, highlighting the importance of distillation for achieving both fast and high-quality video generation. With the same bidirectional teacher, we demonstrate that initializing the student model by fitting the ODE pairs can further enhance performance.
While our student model improves upon the bidirectional teacher for frame-by-frame quality, it performs worse in temporal flickering and output diversity.

\begin{table}[h!]
    \centering
    \scriptsize
    \begin{tabular}{l@{\ \ }c@{\ \ }c@{\ \ }c@{\ \ }c@{\ \ }c@{\ \ }c@{\ \ }c}
    \toprule
    \multicolumn{2}{c}{\multirow{2}{*}{\shortstack[c]{\textit{Many-step models}}}} & \multirow{2}{*}{\shortstack[c]{Causal \\ Generator?}} & \multirow{2}{*}{\shortstack[c]{\# Fwd \\ Pass}} & \multirow{2}{*}{\shortstack[c]{Temporal \\ Quality}} & \multirow{2}{*}{\shortstack[c]{Frame \\ Quality}} & \multirow{2}{*}{\shortstack[c]{Text \\ Alignment}} \\ \\
    \midrule 
    \multicolumn{2}{c}{Bidirectional} & \xmark & 100 & 94.6 & 62.7 & 29.6 \\
    \multicolumn{2}{c}{Causal} & \cmark & 100 & 92.4 & 60.1 & 28.5 \\ 
    \midrule
    \multicolumn{2}{c}{\textit{Few-step models}} & &  &  &  & \\ 
    ODE Init.   & Teacher \\
    \midrule
    \xmark & Bidirectional & \cmark & 4 & 93.4 & 60.6 & 29.4 \\ 
    \cmark           & None          & \cmark & 4 & 92.9 & 48.1 & 25.3 \\ 
    \cmark           & Causal        & \cmark & 4 & 91.9 & 61.7 & 28.2 \\ 
    \cmark           & Bidirectional & \cmark & 4 & \textbf{94.7} & \textbf{64.4} & \textbf{30.1} \\ 
    \bottomrule
    \end{tabular}
    \tablespace
    \caption{\label{table:ablation} Ablation studies. All models generate videos of 10s. The top half presents results of fine-tuning the bidirectional DiT into causal models without few-step distillation. The bottom half compares different design choices in our distillation framework. The last row is our final configuration.}
\end{table}

\subsection{Applications}

In addition to text-to-video generation, our method supports a broad range of other applications. We present quantitative results below, with qualitative samples in Fig.~\ref{fig:main_result}. We provide additional video results in the supplementary material.

\vspace{0.5em}
\noindent\textbf{Streaming Video-to-Video Translation.} We evaluate our method on the task of streaming video-to-video translation, which aims to edit a streaming video input that can have unlimited frames. 
Inspired by SDEdit~\cite{meng2021sdedit}, we inject noise corresponding to timestep $t_1$ into each input video chunk and then denoise it in one step conditioned on the text.
We compare our method with StreamV2V~\cite{liang2024looking}, a state-of-the-art method for this task that builds upon image diffusion models. From 67 video-prompt pairs used in StreamV2V's user study~(originally from the DAVIS~\cite{pont20172017} dataset), we select all 60 videos that contain at least 16 frames. For a fair comparison, we do not apply any concept-specific fine-tuning to either method. Tab.~\ref{table:streamingv2v} shows that our method outperforms StreamV2V, demonstrating improved temporal consistency due to the video prior in our model.

\begin{table}[ht!]
\centering
\scriptsize
\begin{tabular}{l@{\ \ \ }c@{\ \ \ }c@{\ \ \ }c}
\toprule
\multirow{2}{*}{Method} & \multirow{2}{*}{\shortstack[c]{Temporal \\ Quality}} & \multirow{2}{*}{\shortstack[c]{Frame \\ Quality}} & \multirow{2}{*}{\shortstack[c]{Text \\ Alignment}} \\
\\
\midrule 
StreamV2V  &  92.5 &  59.3  & 26.9 \\ 
\textbf{\method~(Ours)} & \textbf{93.2} & \textbf{61.7} &  \textbf{27.7} \\ 
\bottomrule
\end{tabular}
\tablespace
\caption{\label{table:streamingv2v} Evaluation of streaming video-to-video translation.}
\end{table}

\vspace{0.5em}
\noindent\textbf{Image to Video Generation}  
Our model can perform text-conditioned image-to-video generation without any additional training.
Given a text prompt and an initial image, we duplicate the image to create the first segment of frames. The model then autoregressively generates subsequent frames to extend the video.
We achieve compelling results despite the simplicity of this approach.
We evaluate against CogVideoX~\cite{yang2024cogvideox} and Pyramid Flow~\cite{jin2024pyramidal} on the VBench-I2V benchmark, as these are the primary baselines capable of generating 6-10 second videos. As shown in Table~\ref{table:i2v}, our method outperforms existing approaches with notable improvements in dynamic quality.
We believe instruction fine-tuning with a small set of image-to-video data could further enhance our model’s performance.

\begin{table}[ht!]
\centering
\scriptsize
\begin{tabular}{l@{\ \ \ }c@{\ \ \ }c@{\ \ \ }c}
\toprule
\multirow{2}{*}{Method} & \multirow{2}{*}{\shortstack[c]{Temporal \\ Quality}} & \multirow{2}{*}{\shortstack[c]{Frame \\ Quality}} & \multirow{2}{*}{\shortstack[c]{Text \\ Alignment}} \\
\\
\midrule 
CogVideoX-5B  & 87.0 & 64.9 & 28.9     \\ 
Pyramid Flow & 88.4 & 60.3 & 27.6\\
\textbf{\method~(Ours)} & \textbf{92.0} & \textbf{65.0} & \textbf{28.9} \\ 
\bottomrule
\end{tabular}
\tablespace
\caption{\label{table:i2v} Evaluation of image-to-video generation. CogVideoX generates 6s video while the other methods generate 10s video.}
\end{table}

\subsection{Ultra-Long Video Generation}
\begin{figure*}[t]
    \centering
    \includegraphics[width=\linewidth]{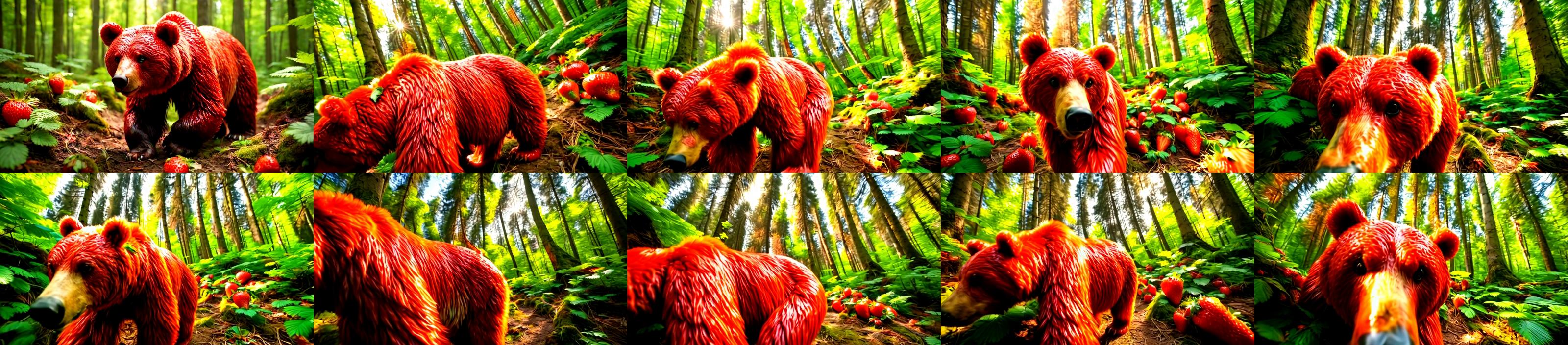} %
    \caption{A 14-minute video example generated by our model. Frames are arranged temporally from left to right and top to bottom.}
    \label{fig:long_video}
\end{figure*}

Our model demonstrates strong performance on videos exceeding 10 minutes in duration. As shown in Fig.~\ref{fig:long_video}, a 14-minute example video exhibits slight overexposure but retains overall high quality.

\section{Discussion}
\label{sec:discussion}
Although our method is able to generate high-quality videos up to 30 seconds, we still observe quality degradation when generating videos that are extremely long.
Developing more effective strategies to address error accumulation remains future work.
Moreover, while the latency is significantly lower—by multiple orders of magnitude—compared to previous approaches, it remains constrained by the current VAE design, which necessitates the generation of five latent frames before producing any output pixels. Adopting a more efficient frame-wise VAE could reduce latency by an additional order of magnitude, significantly improving the model's responsiveness.

Finally, while our method produces high-quality samples using the DMD objective, it comes with reduced output diversity. This limitation is characteristic of reverse KL-based distribution matching approaches. Future work could explore alternative objectives such as EM-Distillation~\cite{xie2024distillation} and Score Implicit Matching~\cite{luo2024one}, which may better preserve the diversity of outputs.

While our current implementation is limited to generating videos at around $10$ FPS, standard engineering optimizations (including model compilation, quantization, and parallelization) could potentially enable real-time performance.
We believe our work marks a significant advancement in video generation and opens up new possibilities for applications in robotic learning~\cite{escontrela2024video,wuunleashing}, game rendering~\cite{valevski2024diffusion,che2024gamegen}, streaming video editing~\cite{chen2024streaming}, and other scenarios that require real-time and long-horizon video generation.

\paragraph{Acknowledgment}
The research was partially supported by the Amazon Science Hub, GIST, Adobe, Google, Quanta Computer, as well as the United States Air Force Research Laboratory and the United States Air Force Artificial Intelligence Accelerator under Cooperative Agreement Number FA8750-19-2-1000. The views and conclusions expressed in this document are those of the authors and do not necessarily represent the official policies or endorsements of the United States Air Force or the U.S. Government. Notwithstanding any copyright statement herein, the U.S. Government is authorized to reproduce and distribute reprints for official purposes.

{
    \small
    \bibliographystyle{ieeenat_fullname}
    \bibliography{main}
}

\clearpage
\maketitlesupplementary

\renewcommand{\thesection}{\Alph{section}}
\setcounter{section}{0}

\section{VBench-Long Leaderboard Results}
\begin{figure*}[ht!]
    \includegraphics[width=\textwidth]{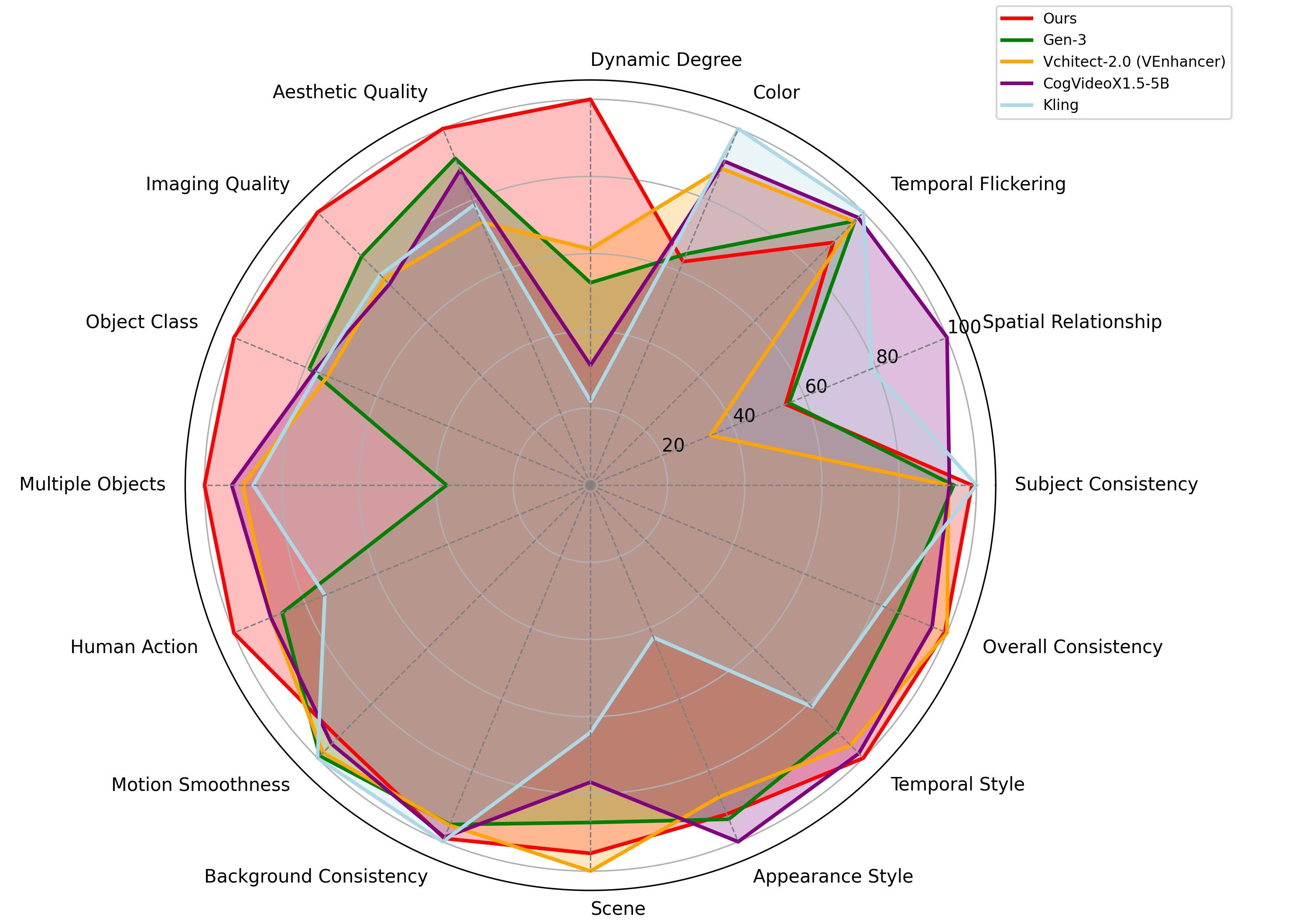}
    \caption{
    VBench metrics and comparison with previous state-of-the-art methods. Our method (red) performs strongly across different dimensions.
    \label{fig:vbench_long} 
    }
\end{figure*}

We evaluate \method~on the VBench-Long dataset using all 946 prompts across 16 standardized metrics. We refer readers to the VBench paper~\cite{huang2024vbench} for a detailed description of the metrics. As shown in Tab.~\ref{tab:vbench_long}, our method achieves state-of-the-art performance with the highest total score of 84.27. The radar plot in Fig.~\ref{fig:vbench_long} visualizes our method's comprehensive performance advantages. Our method is significantly ahead in several key metrics including dynamic degree, aesthetic quality, imaging quality, object class, multiple objects, and human action. More details can be found on the official benchmark website (\url{https://huggingface.co/spaces/Vchitect/VBench_Leaderboard}).

\begin{table*}[hb!]
\centering
\begin{adjustbox}{width=1\textwidth}
\begin{tabular}{lccccccccccccccccccc}
\toprule
\textbf{Method} & \textbf{Total Score} & \textbf{Quality Score} & \textbf{Semantic Score} & \textbf{Subject Consistency} & \textbf{Background Consistency} & \textbf{Temporal Flickering} & \textbf{Motion Smoothness} & \textbf{Dynamic Degree} & \textbf{Aesthetic Quality} & \textbf{Imaging Quality} & \textbf{Object Class} & \textbf{Multiple Objects} & \textbf{Human Action} & \textbf{Color} & \textbf{Spatial Relationship} & \textbf{Scene} & \textbf{Appearance Style} & \textbf{Temporal Style} & \textbf{Overall Consistency} \\
\midrule
Vchitect  & 82.24 & 83.54 & 77.06 & 96.83 & 96.66 & 98.57 & 98.98 & 63.89 & 60.41 & 65.35 & 86.61 & 68.84 & 97.20  & 87.04 & 57.55 & \textbf{56.57} & 23.73 & 25.01 & 27.57 \\
Jimeng    & 81.97 & 83.29 & 76.69 & 97.25 & 98.39 & 99.03 & 98.09 & 38.43 & \textbf{68.80}  & 67.09 & 89.62 & 69.08 & 90.10  & 89.05 & 77.45 & 44.94 & 22.27 & 24.7 & 27.10 \\
CogVideoX & 81.61 & 82.75 & 77.04 & 96.23 & 96.52 & 98.66 & 96.92 & 70.97 & 61.98 & 62.90 & 85.23 & 62.11 & 99.40  & 82.81 & 66.35 & 53.20 & \textbf{24.91} & \textbf{25.38} & \textbf{27.59} \\
Vidu      & 81.89 & 83.85 & 74.04 & 94.63 & 96.55 & 99.08 & 97.71 & 82.64 & 60.87 & 63.32 & 88.43 & 61.68 & 97.40  & 83.24 & 66.18 & 46.07 & 21.54 & 23.79 & 26.47 \\
Kling     & 81.85 & 83.39 & 75.68 & \textbf{98.33} & 97.60 & \textbf{99.30} & \textbf{99.40} & 46.94 & 61.21 & 65.62 & 87.24 & 68.05 & 93.40  & \textbf{89.90}  & 73.03 & 50.86 & 19.62 & 24.17 & 26.42 \\
CogVideoX1.5-5B & 82.17 & 82.78 & \textbf{79.76} & 96.87 & \textbf{97.35} & 98.88 & 98.31 & 50.93 & 62.79 & 65.02 & 87.47 & 69.65 & 97.20  & 87.55 & \textbf{80.25} & 52.91 & 24.89 & 25.19 & 27.30 \\
Gen-3     & 82.32 & 84.11 & 75.17 & 97.10  & 96.62 & 98.61 & 99.23 & 60.14 & 63.34 & 66.82 & 87.81 & 53.64 & 96.40  & 80.90  & 65.09 & 54.57 & 24.31 & 24.71 & 26.69 \\
\textbf{\method~(Ours)}      & \textbf{84.27} & \textbf{85.65} & 78.75  & 97.53 & 97.19 & 96.24 & 98.05 & \textbf{92.69}  & 64.15 & \textbf{68.88}  & \textbf{92.99}  & \textbf{72.15} & \textbf{99.80}  & 80.17 & 64.65 & 56.58 & 24.27 & 25.33 & 27.51 \\
\bottomrule
\end{tabular}
\end{adjustbox}
\caption{Full comparison on VBench-Long using all 16 metrics. The best scores are in bold. Please zoom in for details.}
\label{tab:vbench_long}
\end{table*}

\begin{figure*}[b]
    \centering
    \includegraphics[width=\linewidth]{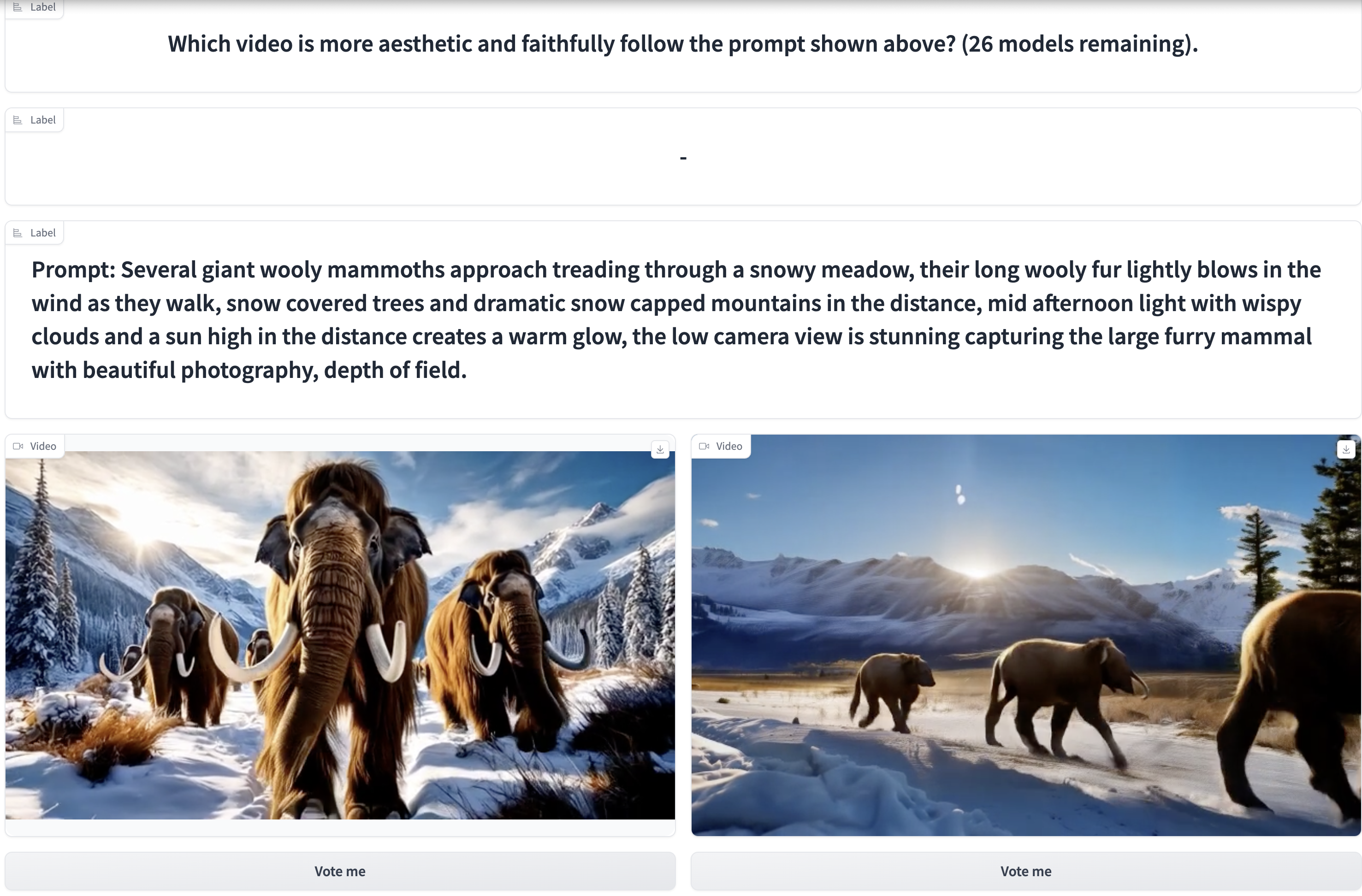}
    \caption{
    An example interface for our user preference study, where videos generated by different methods are displayed in a randomized left/right arrangement.
    \label{fig:interface} 
    }
\end{figure*}

\end{document}